# Multiple Convolutional Neural Network for Skin Dermoscopic Image Classification

Yanhui Guo[1] and Amira S. Ashour[2]

[1]Department of Computer Science, University of Illinois at Springfield, Springfield, Illinois, USA

[2]Department of Electronics and Electrical Communications Engineering, Faculty of Engineering, Tanta University, Egypt

**Abstract**

Melanoma classification is a serious stage to identify the skin disease. It is considered a challenging process due to the intra-class discrepancy of melanomas, skin lesions' low contrast, and the artifacts in the dermoscopy images, including noise, existence of hair, air bubbles, and the similarity between melanoma and non-melanoma cases. To solve these problems, we propose a novel multiple convolution neural network model (MCNN) to classify different seven disease types in dermoscopic images, where several models were trained separately using an additive sample learning strategy. The MCNN model is trained and tested using the training and validation sets from the International Skin Imaging Collaboration (ISIC 2018), respectively. The receiver operating characteristic (ROC) curve is used to evaluate the performance of the proposed method. The values of AUC (the area under the ROC curve) were used to evaluate the performance of the MCNN.

## I. Introduction

Skin abnormalities are the first indicator of different diseases including the skin cancer, which is one of the main health problems worldwide. There are several skin lesion types, including melanoma, basal cell carcinoma, melanocytic nevus, bermatofibroma, benign keratosis, actinic keratosis, and vascular lesion [1]. Early detection, identification and classification of the lesion's type play a significant role in the diagnosis and treatment strategy. Traditionally, the physicians depend on their visual experience to evaluate the patients' lesions case-by-case based on the local lesion patterns compared to the skin in the entire body [2]. However, the automated analysis and classification of the dermoscopic images provide accurate and timeless assessment of the skin lesion type according to the skin surface structures of the lesion regions [3]. Several classification techniques have been applied for skin lesion dermoscopic images classification, such as the artificial neural network (ANN) [4], support vector machine (SVM) [5, 6], bag-of-features model [7], decision tree [8], ensemble classification [9], and the fuzzy logic based classifiers [10]. Compared to statistical methods, the ANN provides reduced error rates. Thus, the ANN is designated and imitated in the computer aided diagnosis (CAD) systems to map a set of features/symptoms to a set of the probable diagnostic classes [11]. However, extra knowledge of the dermoscopic dataset cannot be attained from the neural network, where the parameters of the neural network model are not directly deduced. Additionally, the classification process of the dermoscopy images is still puzzling due to the presence of artifacts, noises, the complexity and variability of the skin lesion structures, illumination variation during the image capturing process, dense hairs, air bubble, and low contrast between normal skin and lesion, which perplex the lesion's border detection, feature extraction and classification processes. Along with the smooth transition in color between the background and the lesion regions that specifies the low contrast between the background and the pigmented lesion.

Consequently, researchers have devoted several classification techniques using deep learning (DL) in different applications. The DL algorithms have the machine learning architecture, which driven with the competency to handle large datasets of complex computations. For image classification, the Convolutional Neural Networks (CNNs) are considered the most prevalent architecture in several applications, especially in the medical image classification processes. The CNN has a significant conceptual framework including weights sharing, local perception of the area, and time domain/ down-sampling space, which guarantee a relative unchanged displacement, distortion and scaling characteristics. For skin lesions classification, few researchers have applied the CNNs to exploit their dominant discrimination ability. There are several CNN designs, such as Inception (GoogleNet), AlexNet, QuocNet, BNInception-V2, and Inception-V3. For dermoscopy images classification, Codella *et al*. [12] employed pre-trained CNNs to extract deep features for further classification of the melanoma and atypical lesions as well as the classification of melanoma and non-melanoma. Kawahara *et al*. [13] applied an AlexNet based fully pre-trained CNN on dermoscopic skin images instead of using CNN from scratch. This CNN was pre-trained on natural image to transfer learning from other domains for reducing the time consuming in the training process with 81.8% classification accuracy. Esteva *et al*. [14] classified 129,450 skin lesion dataset using pre-trained CNN GoogleNet inception-V3 structure. Another study on skin lesion classification implemented a deep residual network using more than 50 layers [15]. This residual network comprised of stacked layers consisted of convolutional layer, rectified linear unit and batch normalization layer. In order to initialize the weights and the de-convolutional layers, a pre-trained network on ImageNet dataset model has been used. The authors increased the dataset by rotating and cropping, major axis alignment and augmentation.

A public challenge on Skin Lesion Analysis Toward Melanoma Detection has been hosted by the International Skin Imaging Collaboration (ISIC) in 2017 as reported in [16] to classify the dermoscopic images into only two classes. Matsunaga *et al*. [17] participated in this challenge (part 3) for skin lesion classification by using an ensemble of CNNs. AdaGrad and RMSProp techniques have been used to fine tune the CNNs. The results depicted that Area Under Curve (AUC) of value 0.953 has been achieved to classify the seborrheic keratosis (SK). However, González-Díaz [18] achieved 0.965 AUC value using the CNN, where several networks have been designed to provide the lesion area identification, lesion segmentation and final diagnosis. In the same challenge, Menegola *et al*. [19] used CNN by employing the ResNet-101 and Inception-v4, which were pre-trained on ImageNet, where 0.943 AUC value has been achieved. These preceding studies established that using pre-trained deep learning approaches to classify dermoscopic images is a promising domain.

This paper applied the proposed model training and testing phases in the classification task of the ISIC2018 dataset, there are 10015, 193, and 1512 images in training, validation and testing sets, respectively. In the ISIC 2018 dataset, there are seven categories including basal cell carcinoma, melanoma, melanocytic nevus, benign keratosis, intraepithelial carcinoma (actinic keratosis/ Bowen's disease), vascular lesion, and dermatofibroma, while the ISIC 2017 skin lesions are diagnosed into two categories: benign and malign. Considering the more challenges in the ISIC2018 classification task, multiple models are necessary to be employed to enhance the performance of individual model methods. At the same time, the multiple model based methods can reduce the training time with fewer epochs on each model.

This work proposed a novel multiple convolution neural network model (MCNN) to classify the dermoscopic images. Transfer learning is used to train multiple models on different samples. The samples are selected according to their scores' performance at the previous models.

The structure of the remaining sections is as follows. In section II, the methodology of the proposed model is introduced. In section III, the proposed novel learning scheme model is included. Afterwards, the simulation results are interpreted and finally the conclusions are presented in sections IV and V; respectively.

## II. Methodology & Proposed Method

In the proposed multiple convolutional neural network (MCNN) model, multiple networks are trained using an additive training strategy to adjust the learning samples for different CNN models. The idea of this strategy is to improve the model's performance by repeated training of the model on wrongly classified samples, which is consistent to the human's recognition procession. In the prediction stage, the classification results of the testing samples are determined by the model which has highest prediction score.

To speed up the training stage, transfer learning is employed, and multiple models are initialized based on a pretrained CNN model, such as AlexNet, VGG16, GoogleNet, and

ResNet. The first network is trained using the whole training set for limited iterations to speed up training. Afterwards, its performance on each sample is described as:

$$T_m(i) = S_m^1(i) \qquad (1)$$

where $S_m^1(i)$ are the highest scores of the sample $i$ for the $m^{th}$ model.

Then, identify the samples whose classification results have poor performance for further use to train the next model repeatedly, where the sample set for the $(m+1)^{th}$ model is given by:

$$ST_{m+1} = \{S_m(i)|T_m(i) < T^*) \qquad (2)$$

where $T^*$ is a threshold which is selected as 0.9 and determined by a trial-and-error method.

This training procedure is applied iteratively until all models are trained using the added samples with poor performance from the previous models.

In testing stage, the input image is classified by a serial of MCNN models and final classification decision is determined according to the model with the highest prediction score, which can be expressed as follows:

$$L(i) = \underset{n}{\mathrm{argmax}}\{S_m^n(i)\} \qquad (3)$$

where $S_m^n(i)$ is the prediction score to category $n$ of the $m^{th}$ model. The final classification result of each sample is obtained from the model with the highest performance score on it.

Accordingly, the proposed MCNN approach can be summarized as follows.

---

***Algorithm: Proposed multiple convolutional neural network***

    ***Step 1: Train*** the first model using the training set for limited iterations
    ***Step 2: Identify*** some wrongly classified samples from the previous training and duplicate samples in the training set of the trained current model
    ***Step 3: Determine*** the samples whose results are less than the threshold and duplicate these samples in the training set
    ***Step 4: Train*** the next model using the additive sample set
    ***Step 5: Repeat*** the previous steps from step 2 until no more wrongly classified and

| |
|---|
| additive samples |
| ***Step 6: Predict*** the samples and determine their classification result according to their performance score of different models |

III. Experimental Results

The available part of the ISIC2018 dataset including 10015 images for training has been used for training, 193 images in validation set, and another 1512 are used to evaluate the proposed MCNN model [20], where this HAM10000 dataset was collected and cleaned by Tschandl *et al*. [21]. The receiver operating characteristic (ROC) curve is used to evaluate the performance of the classifier, which is a useful tool for organizing the classifiers and visualizing their performance. Compared with scalar measures, such as the accuracy, error rate, and the error cost; the ROC graphs can provide richer measure by calculating the AUC (the area under the ROC curve) [22]. The AUC for the training set and validation set are 0.99 and 0.72, respectively.

IV. Conclusion

In this work, a multiple convolutional neural network is proposed to classify seven types of the skin lesion in dermscopic images. The ISIC 2018 dataset is used to train and validate the proposed deep learning model. The achieved results established the effectiveness of the proposed model with AUC as 0.99 and 0.72 on training and validation sets.